\documentclass[letterpaper, 10 pt, conference]{ieeeconf}  
\IEEEoverridecommandlockouts                              
\usepackage{cite}
\usepackage{amsmath, amssymb, amsfonts}
\usepackage{algorithmic}
\usepackage{graphicx}
\usepackage{textcomp}
\usepackage{xcolor}
\usepackage{balance}
\usepackage{gensymb}
\usepackage{subcaption}
\usepackage{multirow}
\def\BibTeX{{\rm B\kern-.05em{\sc i\kern-.025em b}\kern-.08em
    T\kern-.1667em\lower.7ex\hbox{E}\kern-.125emX}}

\usepackage{dblfloatfix} 
\usepackage{notoccite} 

\title{\LARGE \bf
RaccoonBot: An Autonomous Wire-Traversing Solar-Tracking Robot for Persistent Environmental Monitoring
}

\author{Efrain Mendez-Flores$^{1}$, Agaton Pourshahidi$^{2}$ and Magnus Egerstedt$^{3}$
\thanks{*This work was supported by a grant from the Crystal Cove Conservancy.}
\thanks{$^{1}$Efrain Mendez-Flores, $^{2}$Agaton Pourshahidi and $^{3}$Magnus Egerstedt are with the Department of Electrical Engineering and Computer Science, University of California, Irvine, CA 92697, USA. Email: {\tt\small $\{$efrainmf, ajpoursh, magnus$\}$@uci.edu}.}%
}

\begin{document}
\maketitle
\thispagestyle{empty}
\pagestyle{empty}

\begin{abstract}
Environmental monitoring is used to characterize the health and relationship between organisms and their environments. In forest ecosystems, robots can serve as platforms to acquire such data, even in hard-to-reach places where wire-traversing platforms are particularly promising due to their efficient displacement. This paper presents the RaccoonBot, which is a novel autonomous wire-traversing robot for persistent environmental monitoring, featuring a fail-safe mechanical design with a self-locking mechanism in case of electrical shortage. The robot also features energy-aware mobility through a novel Solar tracking algorithm, that allows the robot to find a position on the wire to have direct contact with solar power to increase the energy harvested. Experimental results validate the electro-mechanical features of the RaccoonBot, showing that it is able to handle wire perturbations, different inclinations, and achieving energy autonomy. 
\end{abstract}

\section{Introduction}\label{sec:Intro}

\begin{figure}[h!]
    \centering
    \includegraphics[scale=0.21]{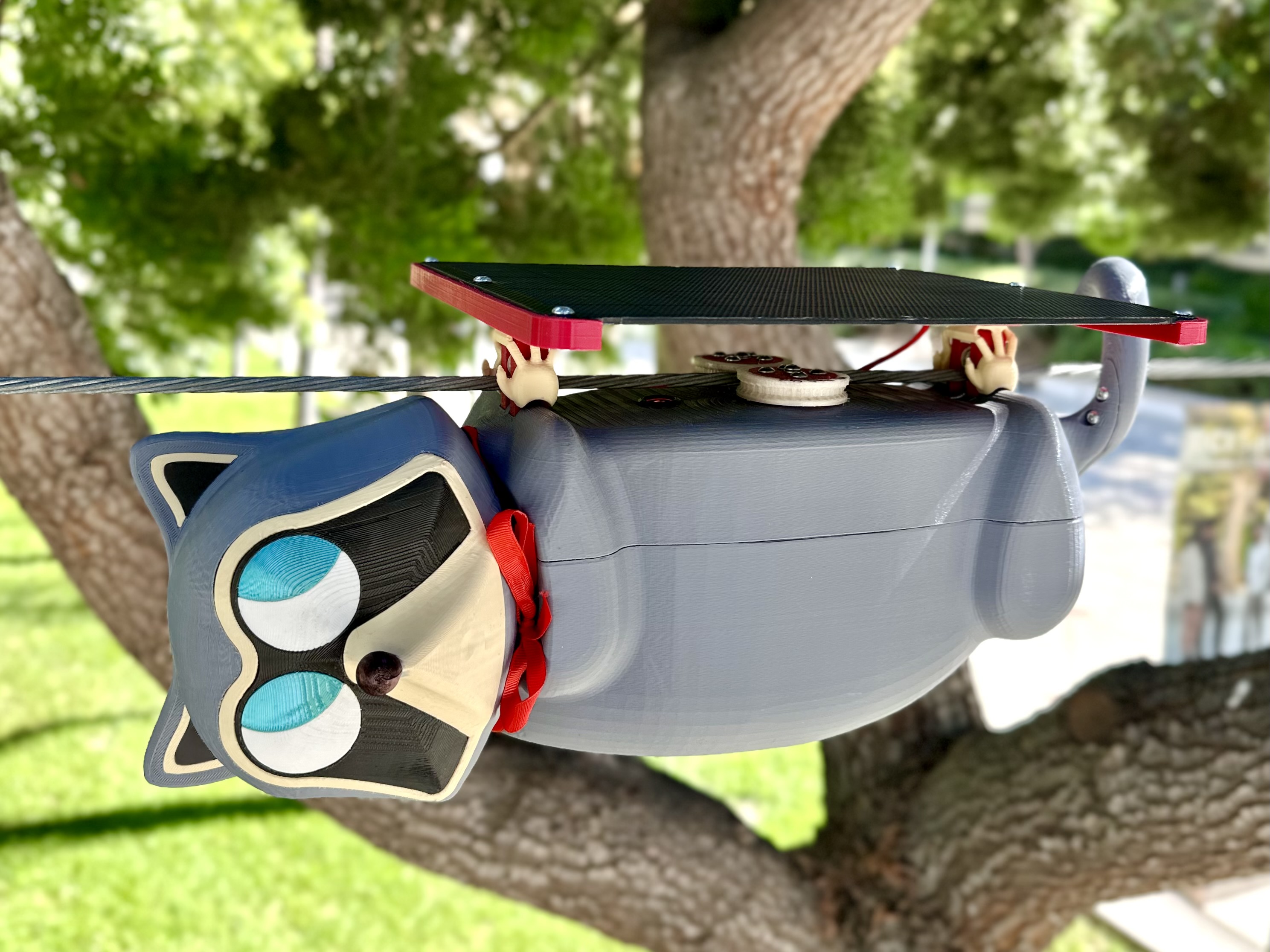}\vspace{-2.5pt}
    \caption{RaccoonBot traveling along a wire between two trees.}
    \label{fig:RaccoonBot}
\end{figure}

The field of ecology concerns itself with the many intricate, intertwined, and complex interconnections formed between different organisms, and between their environments, e.g., \cite{Odum,Ricklefs}. Characterizing such intricate relationships is only possible if sufficient data is available, i.e., if enough meaningful observations can be made about what really goes on in an ecosystem. To that end, environmental monitoring involves the gathering of data in some environment of interest. But, it becomes useful only in some broader context in which the data is actionable. Conservation biology is a prime example of such a context, as it is a discipline focused on the management of nature and of Earth's biodiversity, \cite{primack1993essentials}. To support this aim, this paper presents the RaccoonBot (Fig.~\ref{fig:RaccoonBot}), a novel environmental monitoring robot, particularly suited for forest ecosystems.

\vspace{-5pt}
\vspace{-7.5pt}

Autonomous robotic systems are particularly well-suited for environmental monitoring due to their ability to navigate difficult or even inaccessible terrain, while ensuring continuous 24/7 operation without the need for human intervention. Different types of robots for environmental preservation and monitoring of forests are compared in \cite{oliveira2021advances}, where it is clear that the high irregularities of the soil complicate the application of ground robots, and opens the door for less invasive locomotions; such as, UAV's or climbing robots.

Aerial robots offer promising results due to their mobility and extended range. For example, \cite{zhou2022swarm} presents a swarm of micro-drones flying through a dense forest environment, proving their suitability for such environments. However, from a persistent monitoring perspective, these solutions are costly from an energy-consumption vantage-point, requiring frequent recharging or refueling, which is impractical for deployments lasting weeks, months, or even years \cite{kirchgeorg2023design}. While fixed-wing or dirigible aerial robots may be more energy-efficient, their maneuverability is simply not up to the task.

In contrast, \cite{kirchgeorg2022multimodal, kirchgeorg2023design} present a multimodal robot that combines aerial and tethered locomotion, with a design that increases operating times through improved mechanical features. Yet, robots relying on propellers still struggle under windy conditions. Although \cite{kirchgeorg2023design} shows significant improvements in energy density compared to UAVs, the robot is still limited to one battery charge cycle, which is a limitation for long-term environmental monitoring applications \cite{oliveira2021advances}.

\begin{figure*}[hb!]
    \centering
    \includegraphics[scale=0.311]{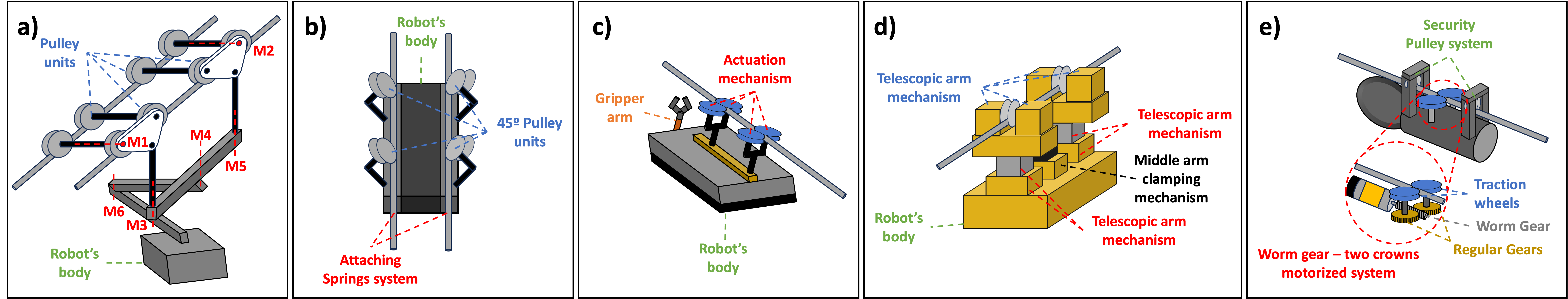}
    \caption{\textbf{a)}~Typical frame of Robots for live line inspections. \textbf{b)}~Mechanical representation of the LineRanger design, \cite{live4_ino}. \textbf{c)}~Robot schematic of a single line inspection robot with two pairs of mechanized wheels, \cite{bahrami2016novel}. \textbf{d)}~Mechanical representation of a single line inspection robot, \cite{singleLineModel}. \textbf{e)}~Mechanical features sketch of the proposed robot's wire-traversing mechanism.}
    \label{fig:Robot_all}
\end{figure*}

So if the robots cannot be on the ground, nor in the air, where should they be? Well, they could climb. This is what arboreal animals do after all, so why not let the robots follow their example? Climbing robots would indeed be an elegant solution. But such robots run the real risk of falling out of the trees, as explored in \cite{kirchgeorg2023design}. One option could be to slightly engineer the environment so that the robots can stay close to the strategies employed by climbing, arboreal animals, yet remain safely suspended in the tree canopies for long periods of time. For these reasons, we choose to approach the question of how to perform environmental monitoring\index{environment!monitoring} tasks in the tree canopies by letting the robots be wire-traversing, i.e., having them move along wires mounted between trees.

Although wire-traversing robots for environmental monitoring purposes have been previously considered, e.g., \cite{notomista2019slothbot,smith2012persistent}, the novelty in this paper lies in the fail-safe mechanical design, and the efficient power electronics, which includes the actuation and power supply for the robot, integrated with a novel solar tracking algorithm for mobile robots. In particular, the RaccoonBot is designed to be reliable against severe wire perturbations, to be capable of dealing with rugged environmental conditions. Achieved through a fail-safe mechanism that even without power, is capable of keeping its position due to its non-reversibility features. This means that without a control signal, the robot will not allow any displacement, even against external perturbations (such as wind or objects colliding with the robot).

Since finding the Maximum Power Point (MPP) in Photovoltaic (PV) systems is an unpredictable dynamic optimization issue (since the MPP constantly changes with irradiation and temperature variations), research has been conducted to improve Maximum Power Point Tracking (MPPT) algorithms for power electronic systems through DC-DC converters~\cite{femia2017power}. But, since the position and orientation of solar panels affect the energy harvesting, Solar Tracking Systems (STS) have been developed mostly focused on static solar panels' orientation, e.g., \cite{kumba2022performance, palomino2023optimal}. Yet, the RaccoonBot needs an algorithm to find the best position on the wire to even have access to solar power. This paper contributes a novel solar tracking algorithm, enabling energy-aware mobility for the robot beyond a single battery charge.

This paper is organized as follows: Section~\ref{sec:Cable} compares the state-of-the-art mechanisms with the proposed RaccoonBot's mechanical features. Section~\ref{sec:RB} describes the RaccoonBot's electro-mechanical components. Section~\ref{sec:SolarTrack} introduces the solar tracking algorithm for PV power harvesting. The results of the validation tests are discussed in Section~\ref{sec:results}, and the conclusions are presented in Section~\ref{sec:Conclusions}.

\section{Wire-Traversing Robots}\label{sec:Cable}

{W}{ire}-traversing robots are basically any types of robots capable of moving along cables, wires, and similar structures, \cite{notomista2019slothbot,toussaint2009transmission}, which makes them suitable for applications beyond environmental monitoring\index{environment!monitoring}; such as, agricultural robotics \cite{billingsley2008robotics}, or maintenance in hazardous locations, as in the case of power line inspection, e.g., \cite{pouliot2008geometric}. In fact, this latter application has been a catalyst for different wire-traversing robot styles,\index{wire-traversing!robot} e.g., \cite{nayyerloo2007mechanical,phillips2014line,sawada1991mobile}. Across the wire-traversing design spectrum, the common features include simplicity of the overall system design and, consequently, of motion control, relatively small localization errors, navigation complexity, and low energy needs,~\cite{nayyerloo2009cable,toussaint2009transmission}.

\subsection{Design Considerations: Previous Work}

When designing a compact and reliable robotic platform that ensures safe attachment to the wire, there are several examples to draw inspiration from - many of which have been developed for inspection of power transmission lines, \cite{ekren2023review}. A common mechanism used by robots on live lines is shown in Fig.~\ref{fig:Robot_all}a, which highlights how pulleys and motors are arranged to provide a safe displacement for inspection robots (eg., \cite{live1}, \cite{live2} \& \cite{live3}). The main mechanical features of this type of robots, rely on motors above the transmission line for the displacement and on the pulley units for stability. This design ensures steady displacement due to the vertical traction from the pulleys and motors. However, since the movement is along the horizontal axis, the actuation is significantly impacted by vertical loads.

A modification to this design was introduced in \cite{live4_ino}, where the main innovation was introducing arms connected to the robot base via springs (Fig.~\ref{fig:Robot_all}b). These springs regulate the pressure applied to the wires, acting as both actuation points and safety locks to ensure a reliable operation under different conditions. While the springs enable a safe design, they also increase the motor's consumption due to the added pressure. Also, these robots use two wires for their locomotion, but a minimally invasive single-wire mechanism is preferred for environmental monitoring. 

Single-wire mechanisms add the need for fail-safe mechanical designs; since wind, rain and wire perturbations represent higher risks the robot. An example is presented in \cite{qing2016mechanical} and shown in Fig.~ \ref{fig:Robot_all}c, where two pairs of motorized pulleys are placed to enhance a smooth displacement, that, as per \cite{ekren2023review}, features a prototype with improved dynamic stability due to the grip mechanism of the traction wheels. 

Fig.~\ref{fig:Robot_all}d, shows a different approach for single wire-traversing robots, which rely on actuators on top of the rope linked to the pulleys for the traversing mechanism of this robot. Some of these robots include safety features; such as, palming enclosures to enhance the robustness against different conditions (eg. \cite{singleLineModel} \& \cite{qing2016mechanical}), but some others only rely on the friction of the attachment pulleys to remain in the rope, without any additional component that ensures resisting weather perturbations (eg. \cite{cao2020design}, \cite{rigatos2022nonlinear} \& \cite{yue2022automatic}). The SlothBot \cite{notomista2019slothbot}, which served as a starting point for the RaccoonBot discussed in this paper, is included among these robots.

\subsection{Design Choices}

In light of these previous design considerations, the mechanism selected for the robot proposed in this paper (Fig.~\ref{fig:Robot_all}e) keeps the most desirable features of these prototypes. This is primarily done to ensure that the robot will safely remain attached to the wire even under severe weather conditions. 

The robot design in Fig.~\ref{fig:Robot_all}e features a pulley at each end, similar to the locking mechanism in \cite{singleLineModel}. Yet, unlike that design, which uses a motorized system at each end, this proposal has no such thing. Instead, the pulleys serve as a safety feature, acting both as locking and force decoupling systems; since, when calculating the sum of forces in the actuator, the vertical components are nullified since both ends carry the robot's weight , while the motorized system in the center only rely on friction for the displacement.

The horizontal force components at the support points are canceled since the pulleys at each end act as bearings. This allows an efficient and smooth displacement along the wire, as the motor torque is better exploited by the worm-gear mechanism with a single worm and two crowns. The mechanism is non-reversible, which means that the wheels can only move when torque is applied by the actuator, but the mechanism is not able to move if torque is transmitted from the gears to the worm, which locks the mechanism without the need of an external force. That enhances the reliability of the robot to be safe even when it looses power.

\section{RaccoonBot's Hardware}\label{sec:RB}

The RaccoonBot is a mechatronic system integrated by the gearbox design, an efficient electrical design, and an energy harvesting algorithm to ensure persistent environmental monitoring capabilities. 

\subsection{Mechanical Features}

\begin{figure}[hbt!]
    \centering
    \includegraphics[scale=0.17]{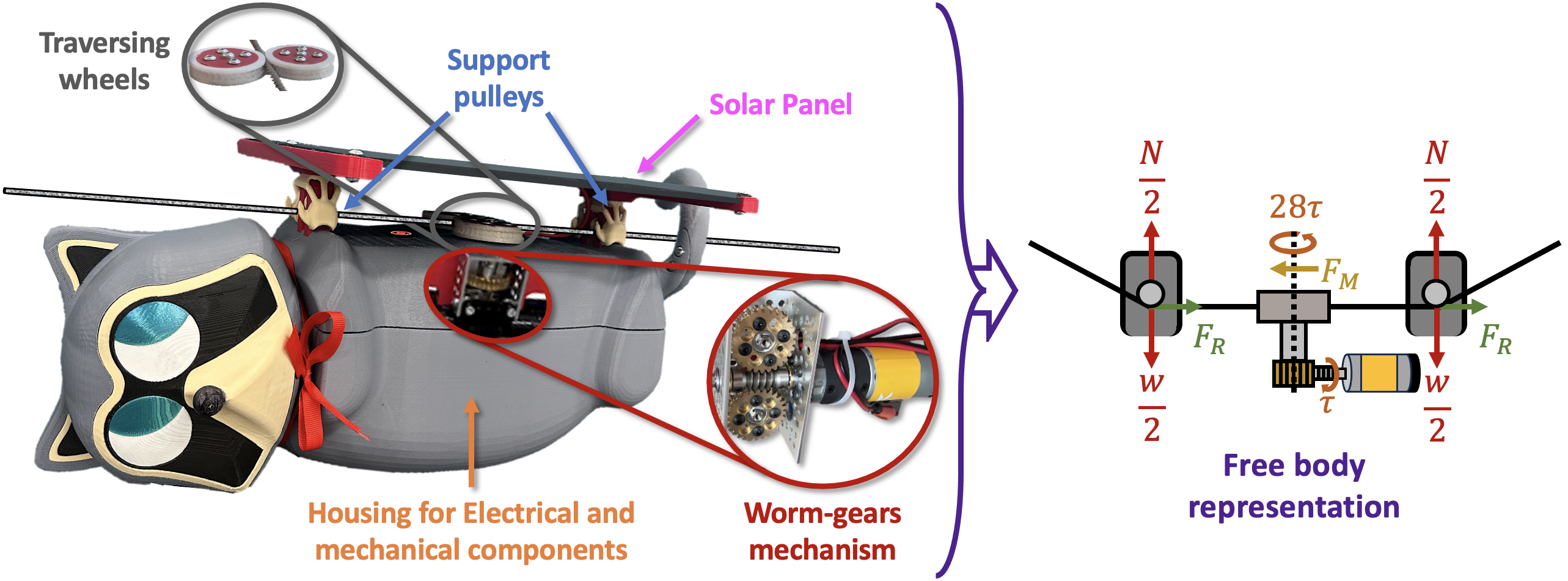}\vspace{-5pt}
    \caption{Mechanical features of the RaccoonBot.}
    \label{fig:MechFeats}
\end{figure}

Among the objectives to be satisfied, reliability against wire perturbations to ensure a fail-safe design is critical for environmental monitoring, where the robot must remain on the wire even in the event of a sudden electrical failure. With those considerations in mind, Fig.~\ref{fig:MechFeats} shows an overview of the mechanical features of the robot.

\vspace{-5pt}
\vspace{-10pt}

The robot relies on a combination of support pulleys and a wire-traversing mechanism, as shown in Fig.~\ref{fig:MechFeats}, where the worm-gear mechanism housed inside the RaccoonBot's body. In fact, the wheels are explicitly designed to support the robot's weight and can move along the wire without pulleys due to their conical shape. However, the pulley arrangement enables force-decomposition, since the pulleys, placed as the robot's "hands", handle the vertical forces caused by the weight, leaving the actuation mechanism affected only by horizontal forces, as shown in Fig.~\ref{fig:MechFeats}'s Free-body diagram.

In Fig.~\ref{fig:MechFeats}, the conversion ratio of the torque ($\tau$), given by the motor, is shown to be 28:1 in the gears due to the gearbox reduction, which is later translated into the force in which the robot is moving along the wire (Force induced by the motor $F_M$). On the other hand, the friction force ($F_g$) opposing the displacement occurs mostly in the support pulleys. Also, Fig.~\ref{fig:MechFeats} shows how the weight and the resultant normal forces balance themselves in the supports, without affecting the central traversing mechanism, yet tensing the wire between them to also smooth the path for the wheels, which, in turn, improve the effects of the motor's torque.

\subsection{Electrical Features}
\begin{figure}[h!]
    \centering
    \includegraphics[scale=0.26]{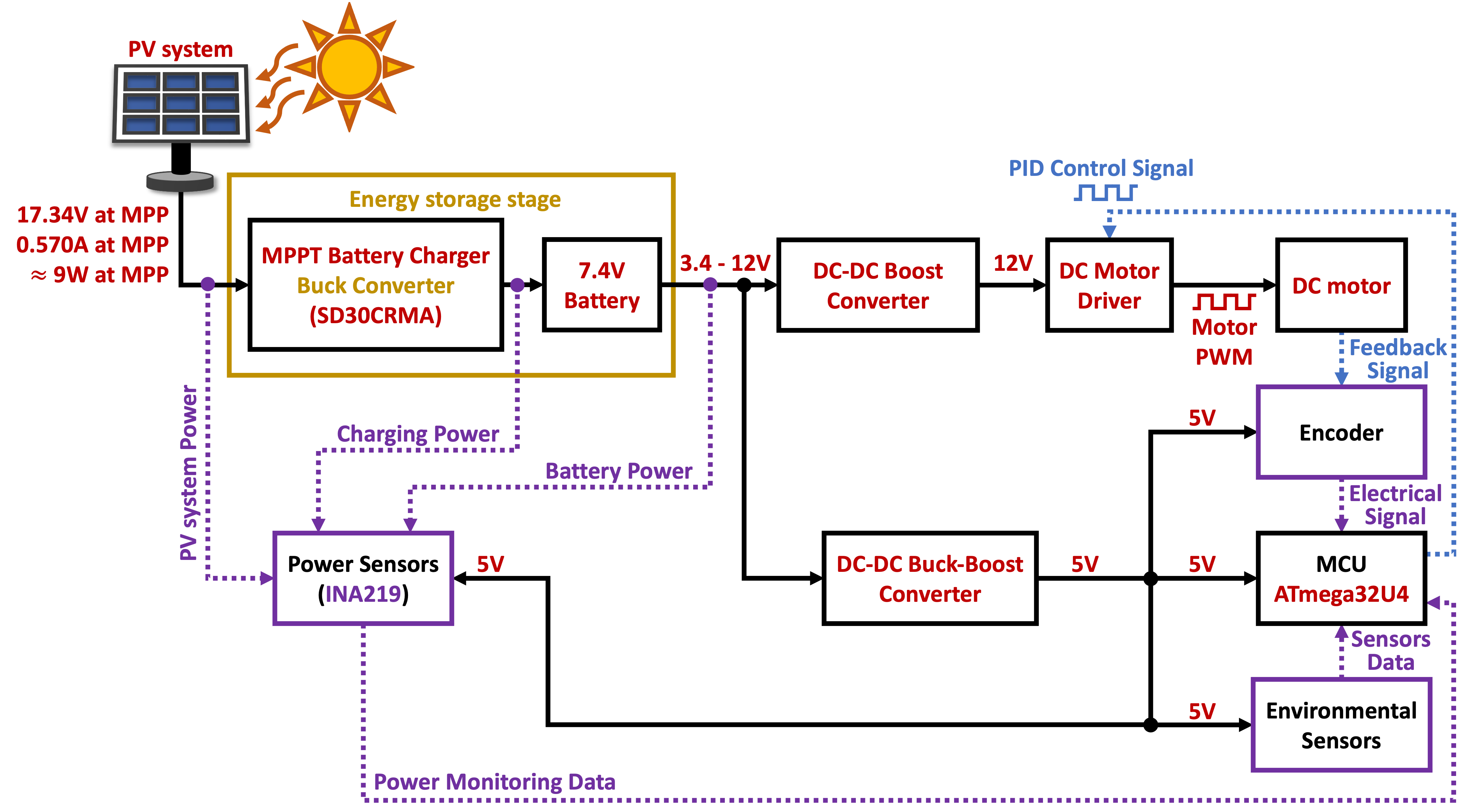}\vspace{-5pt}
    \caption{Electrical structure embedded inside the RaccoonBot.}
    \label{fig:ElecFeats}
\end{figure}

For the RaccoonBot, a reliable and efficient electronics design is critical to achieve the desired, sustainable electrical autonomy. Fig.~\ref{fig:ElecFeats} presents the blocks-diagram with the most relevant components of the electrical structure.

\vspace{-10pt}
\vspace{-10pt}

The energy storage stage in Fig~\ref{fig:ElecFeats}, contains the main circuits integrated to convert photovoltaic energy into the electrical power that keeps the battery charged, which is later used to power all the embedded electrical components, where two DC-DC converters are in charge of regulating the power inside the robot due to the well known efficiency of switching power supplies, when compared to linear regulators.

A boost converter steps-up the battery voltage to the required 12V for the DC motor driver (H-bridge), allowing the motor to operate even when the battery voltage drops from 8.2V to 4V. On the other hand, a buck-boost converter provides a stable 5V output for all the robot's logic components. Both circuits have an average efficiency around 90\%, minimizing energy losses from parasitic components.

\section{Solar Tracking Algorithm}\label{sec:SolarTrack}

\begin{figure}[h!]
    \centering
    \includegraphics[scale=0.25]{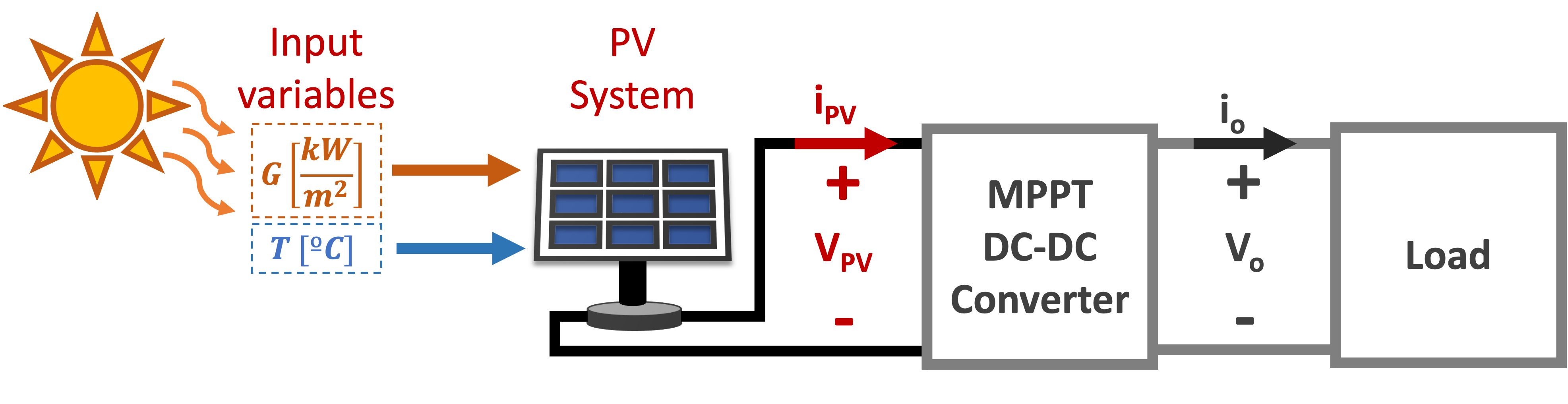}\vspace{-5pt}
    \caption{Simple MPPT structure to regulate the voltage ($V_{PV}$) and current ($i_{PV}$) provided by the solar panel.}
    \label{fig:MPPT}
\end{figure}

The Temperature ($T$) and Solar irradiation ($G$) variations during a regular day, make energy harvesting from PV sources an unpredictable dynamic optimization issue~\cite{femia2017power}; which from the power electronics perspective, is usually dealt through MPPT algorithms implemented through DC-DC converters (as exemplified in Fig~\ref{fig:MPPT}).


\vspace{-10pt}
\vspace{-7.5pt}

Also the orientation and position of the solar panels have a critical role in the energy harvesting process, thereby Solar Tracking Systems (STS) have gained increasing interest~\cite{palomino2023optimal}. STS research has been particularly focused on the orientation of fixed solar panels; as analyzed in~\cite{kumba2022performance} where a control structure is implemented for a Single-Axis STS, and~\cite{palomino2023optimal} where multiple Dual-Axis STS are analyzed.

\begin{figure}[h!]
    \centering
    \includegraphics[scale=0.52]{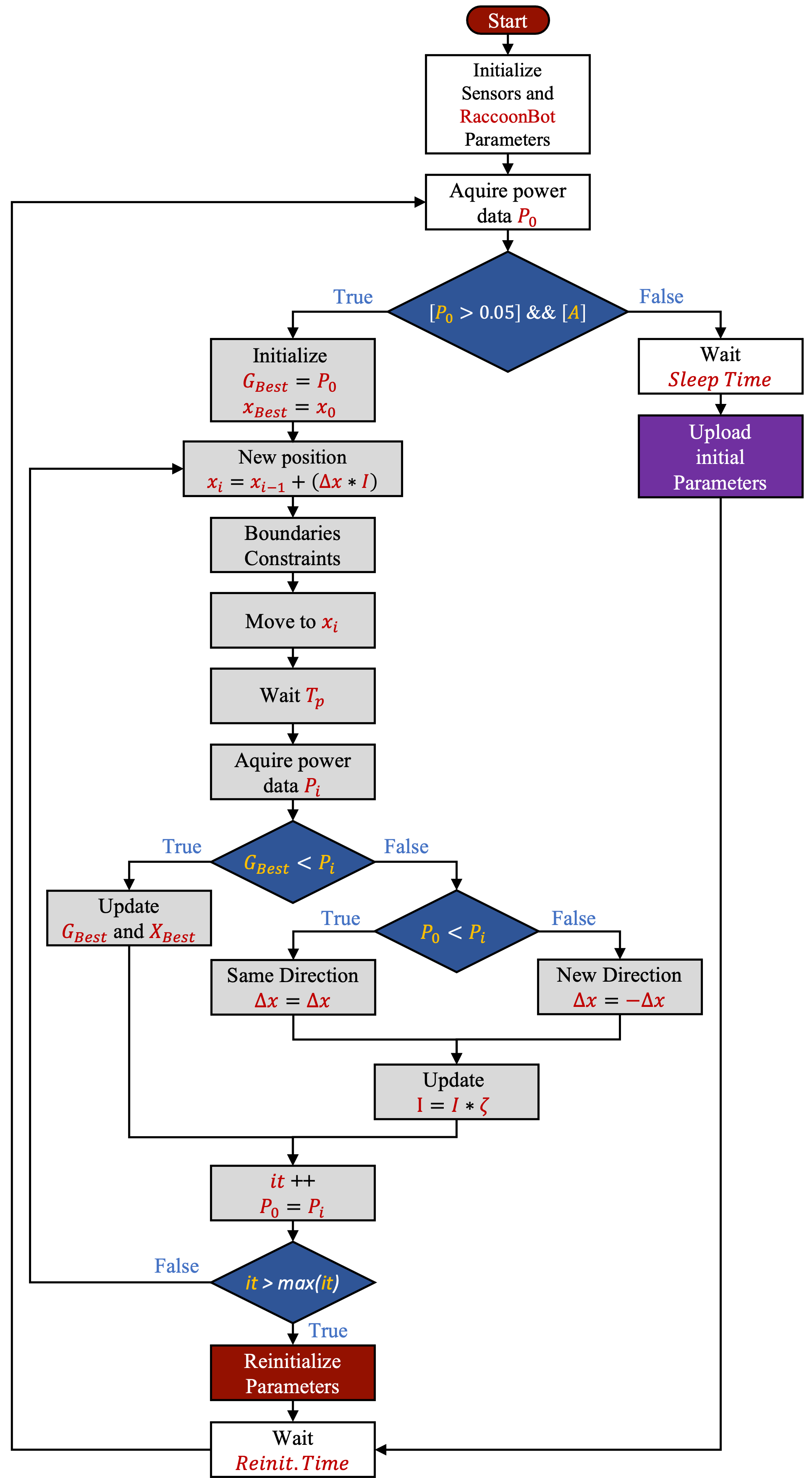}\vspace{-2.5pt}
    \caption{Flowchart diagram of the STS algorithm.}
    \label{fig:RaccoonBotAlg}
\end{figure}

Yet, fixed Solar Tracking Systems (STS) are not suitable for the RaccoonBot due to natural tree shading and the solar dynamics during the day. The robot has to be able to perform as a mobile STS, moving along the wire to find a position with direct access to PV power. This paper, develops a novel mobile STS algorithm (Fig.~\ref{fig:RaccoonBotAlg}), suitable for wire-traversing robots or those needing to find an optimal position along a horizontal line to harness sun power.

\vspace{-11pt}
\vspace{-7.5pt}

The flowchart from Fig.~\ref{fig:RaccoonBotAlg}, shows the general structure of proposed the algorithm; where, the algorithm starts with the initialization of the sensors, microcontroller peripherals and parameters. In this STS application, INA219 power sensors are the ICs selected to carry the power measurements due to their reliability and high accuracy~\cite{bradley2021electrical}.

The first step is to acquire the initial power measurement ($P_0$) and decide whether the data represents that the power from the solar panel means that there is available sun power around or not. On the given configuration, when there is sunlight available the solar panel measurement gives a value around 0.07~[W] under the shade of the trees, since there is still some solar radiation. Hence, the threshold for the algorithm is selected to be grater than 0.05~[W], to decide if the robot should start looking for a new $G_{Best}$.

The second condition to meet to enter the main STS algorithm, is shown in Fig.~\ref{fig:RaccoonBotAlg} as the $A$ condition, which is a change of variable to address the conditions given by~\eqref{eq:Irra}.
\begin{equation}
    A = (P_0 < (G_{Best}-G_{Best}*0.2))\text{or}(G_{Best} == 0)\label{eq:Irra}
\end{equation}where $G_{Best}$ is the global best power measurement acquired at the global best position $x_{Best}$ in the wire. On the left side of the or, the inequality represents how the robot decides when is it necessary to start looking again for a new best position, which is when the robot detects that the acquired power ($P_0$) is 20\% less than the saved $G_{Best}$ power. 

\begin{figure}[h!]
    \centering
    \includegraphics[scale=0.475]{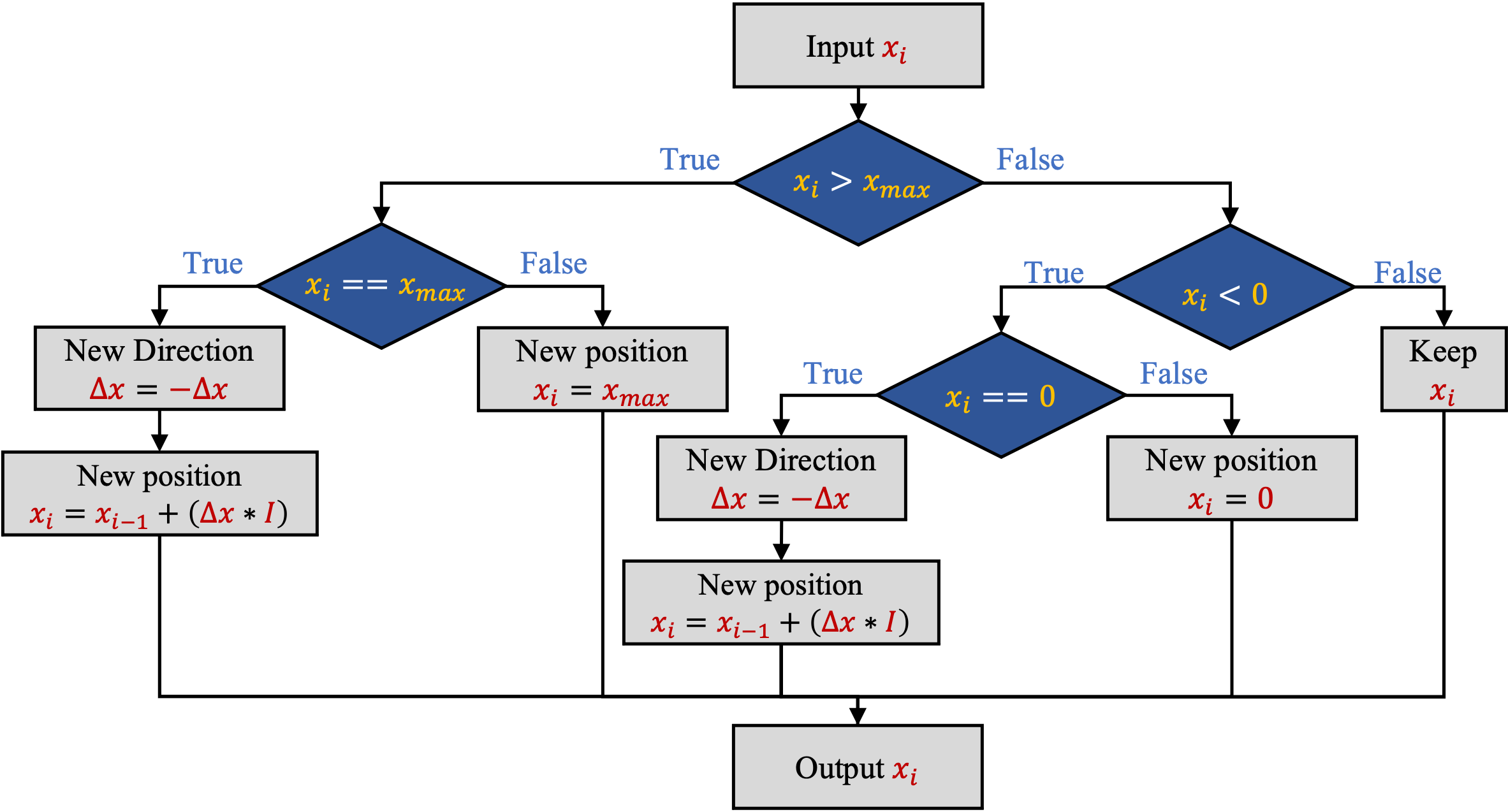}\vspace{-2.5pt}
    \caption{Flowchart diagram behind the boundaries constraints block from Fig.~\ref{fig:RaccoonBotAlg}.}
    \label{fig:boundaries}
\end{figure}

The second half of \eqref{eq:Irra}, represent the initial conditions that enables the algorithm to enter at least once, when the $G_{Best}$ power is initialized as $G_{Best}=0$. Therefore, when the first condition is false, the RaccoonBot will be remain in "sleep mode" and the algorithm will be reinitialized with the initial conditions. When the condition is true, the main optimization process (highlighted in gray from Fig.~\ref{fig:RaccoonBotAlg}) is performed; where, the first step is to take as initial values the acquired power at the current position as the best ones ($G_{Best}$ and $x_{Best}$). Then, the following step is to estimate the next position of the robot at iteration $i$ through \eqref{eq:x_pos}.
\begin{equation}
    x_{i} = x_{i-1} + (\Delta x * I)
    \label{eq:x_pos}
\end{equation}where $x_{i}$ is the new position of the robot, $x_{i-1}$ is the actual position of the robot in the wire, $\Delta x$ is the step size that the robot is going to take at each iteration, and $I$ is the inertia coefficient that enables the convergence through time. After the estimation of $x_{i}$, the result is constrained to the length of the wire where the robot is installed on. First, the algorithm must ensure that $x_{i}$ is within the limits described by Fig.~\ref{fig:boundaries}.

\vspace{-10pt}
\vspace{-7.5pt}

After constraining $x_{i}$, the robot moves to the desired position, and subsequently waits for a period of time $T_p$ to ensure that the MPPT had found the MPP at the $x_{i}$ position. This is a standard in devices that involve MPPT circuits, since \cite{deboucha2023ultra} and \cite{femia2017power} validate that MPPT algorithms can take up to 50~[mSec] since most of them are based on the classic Perturb and Observe (P\&O) algorithm; consequently, $T_p$ should be $\geq$ 50~[mSec] to ensure that the MPPT circuit had enough time to reach the MPP. 

After waiting for a $T_p$ period, the new power value $P_{i}$ is acquired. With that new value, $P_{i}$ and $G_{Best}$ are compared to rank the power values and update $G_{Best}$ and $x_{Best}$ in the case of a new global best. On the other hand, if $P_{i}$ is not a new $G_{Best}$, then the algorithm takes the decision of what direction to take next, to keep looking for the optimal position based on the vector generated by $P_{0}$ and $P_{i}$.

The decision of the new direction, also leads to update the inertia coefficient $I$ through \eqref{eq:I} for the next iteration, since $I$ is the term in charge of the convergence of the algorithm.
\begin{equation}
    I = I * \zeta
    \label{eq:I}
\end{equation}where $\zeta$ is the damping coefficient, which represents the rate of change of the step size. Finally, the algorithm $P_0$ is updated and the decision of whether a new iteration is needed or not is taken.

The difference between the purple and the red reinitialization blocks from Fig.~\ref{fig:RaccoonBotAlg}, is mostly the step size $\Delta x$; since under initial conditions $\Delta x$ is taken as a bigger step size compared to the reinitialization from the red block. Starting with a smaller $\Delta x$, reduces the searching area of the robot and the probability of finding MPPs at the ends of the wire.

At the beginning, a wider search is needed to map the whole area, and to place the robot in a better position for the following optimization trials. But, re-initializing the algorithm with a smaller $\Delta x$ enables a faster, more precise and hence with less energy consumption solar tracking behavior.

\begin{figure}[h!]
    \centering
    \includegraphics[scale=0.241]{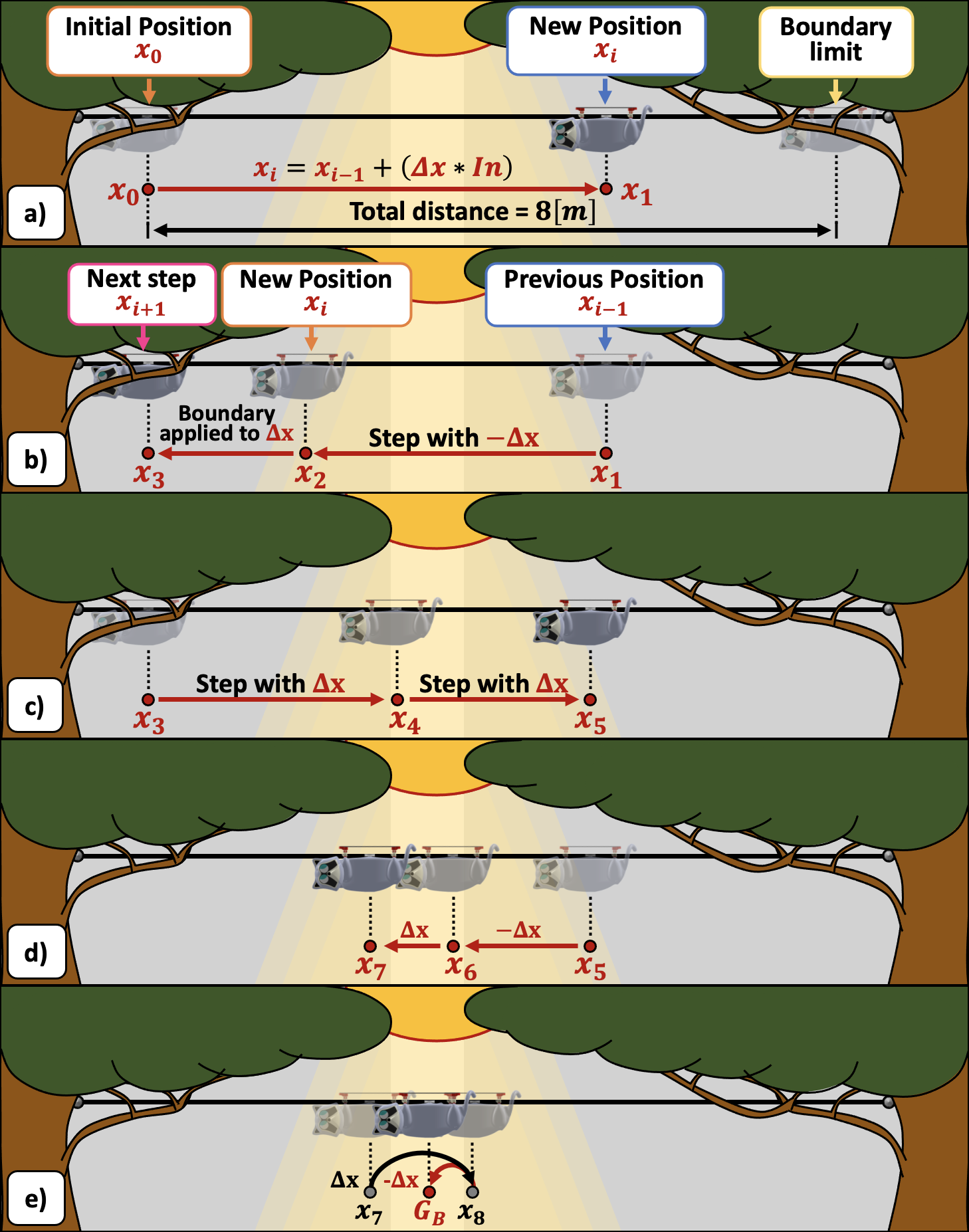}\vspace{-5pt}
    \caption{RaccoonBot's algorithm dynamics.}
    \label{fig:AlgorithmFigs}
\end{figure}

Fig.~\ref{fig:AlgorithmFigs} exemplifies the dynamics of the robot during the implementation of the algorithm, showing a sequence on how the robot decides where to look next for the possible best position until it converges into $G_{Best}$.

\vspace{-7.5pt}
\vspace{-12pt}

\section{Experimental Results}\label{sec:results}

\begin{figure}[h!]
    \centering
    \includegraphics[scale=0.21]{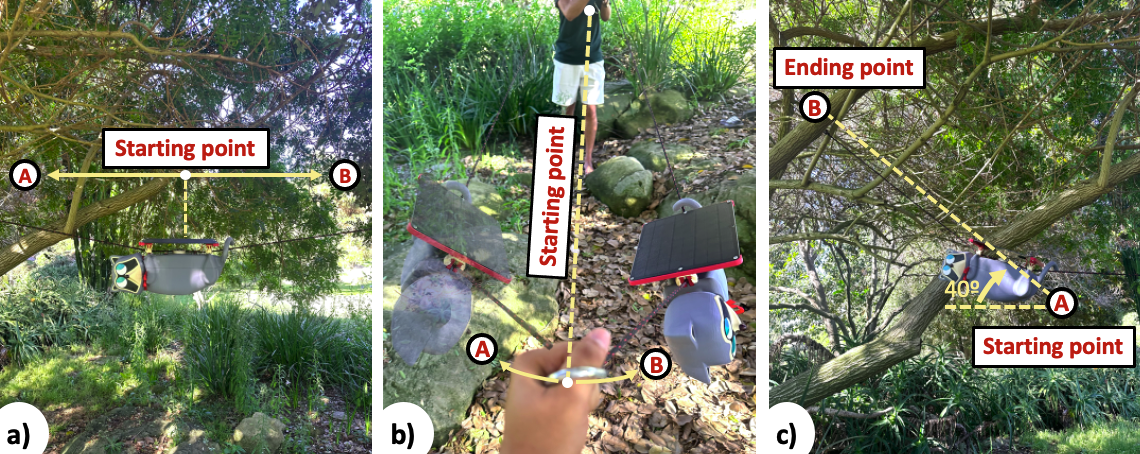}\vspace{-2.5pt}
    \caption{Frames from the: \textbf{a)}~Mobility and Battery durability test, \textbf{b)}~Wire disturbances test, and \textbf{c)}~Climbing test.}
    \label{fig:Test4}
\end{figure}

Since this paper contributes the design of a fail-safe robotic platform for long-term environmental monitoring, it's important to verify that this goal is achieved. To that end, a series of validation tests were conducted. Here, we discuss the outcomes of these tests. Fig.~\ref{fig:Test4} shows frames from the experimental tests in a natural environment.

\vspace{-5pt}
\vspace{-5pt}

Fig.~\ref{fig:Test4}a captures a moment from the mobility and battery durability test, where the robot completed at least 10 loops between points A and B, successfully validating the mechanism's smooth transition and displacement along the wire. On the other hand, Fig.~\ref{fig:Test4}b shows two superimposed frames from the wire disturbance test, demonstrating the robot's ability to stay on a slack climbing rope and continue climbing even under significant centripetal accelerations induced by swinging the rope between points A and B.

The frame in Fig.~\ref{fig:Test4}c shows the robot's ability to climb a 40º slope without any issue, thanks to the PID controller and the non-reversibility features that prevent sliding, even during electrical failures. This test also validates the robot's capability to regulate the motor's torque to achieve a constant speed of 1 meter per minute during both ascent and descent, which represents less than 50\% of its maximum available speed, but still enabling a proper balance between power consumption and torque.

\begin{figure}[h!]
    \centering
    \includegraphics[scale=0.435]{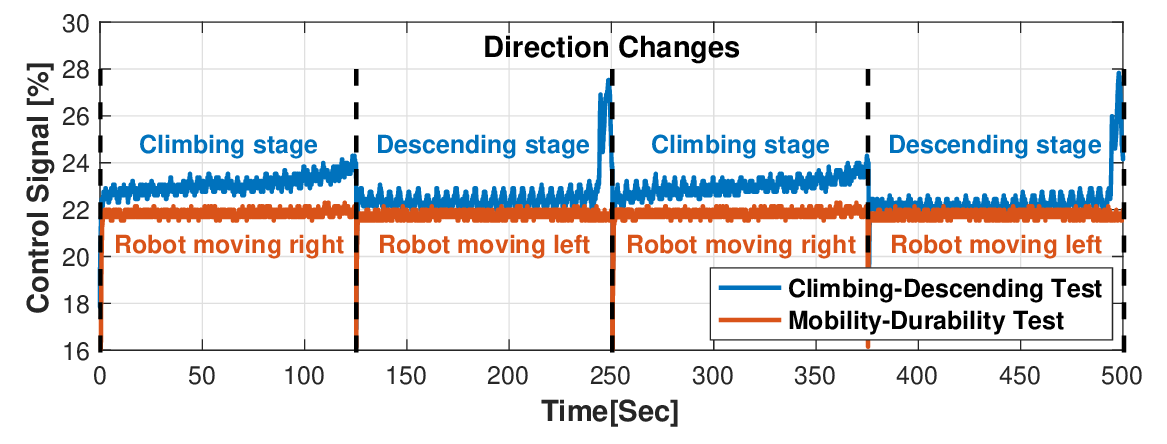}\vspace{-10pt}
    \caption{Control signals from the Mobility test and the Climbing test under multiple repetitions.}
    \label{fig:controlSignals3}
\end{figure}

To illustrate the RaccoonBot's dynamic behavior during the validation tests, Fig.~\ref{fig:controlSignals3} shows the control signals ([\%] of duty cycle) as the robot moves from end to end and returns. The orange profile shows the smooth transitions when the robot travels between points A and B from Fig.~\ref{fig:Test4}a.

\vspace{-10pt}
\vspace{-7.5pt}

The blue profile compares the robot's performance when climbing or descending the slope addressed in Fig.~\ref{fig:Test4}c. Thus, the control signals demonstrate how the robot adapts by increasing power as the slope steepens due to rope tension and adjusting inertia when switching from the descending to the climbing stages. Additionally, Fig.~\ref{fig:controlSignals3} further confirms that, no matter the direction or the slope the robot faces, the RaccoonBot maintains a consistent dynamic behavior across multiple repetitions, validating the stability of its control signals and hence the torque dynamics through time.

Fig.~\ref{fig:controlSignals3} also validates the non-reversible features of the mechanism, since it is clear to see that the control signals from the descending stage closely resemble the mobility test profiles. That happens since when no torque is applied by the motor, the tires are not sliding and the torque generated by the gravity and the weight of the robot is not transmitted to the worm gear. Instead, the motor's torque directly generates the inertia to enable the displacement.

\begin{figure}[h!]
    \centering
    \includegraphics[scale=0.435]{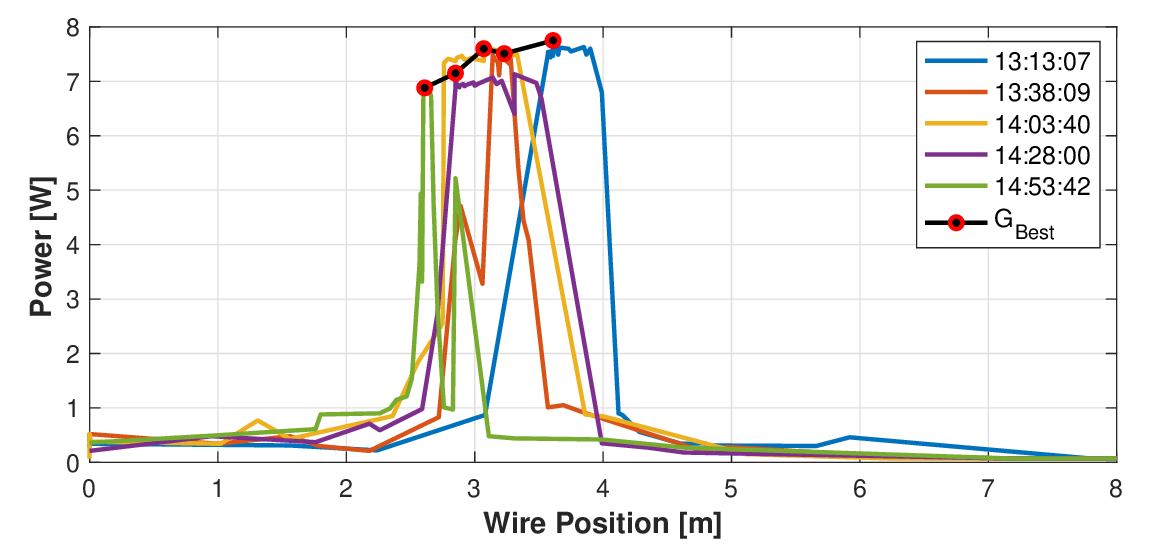}\vspace{-10pt}
    \caption{Power data acquired during 5 solar tracking trials with each $G_{Best}$ highlighted in red.}
    \label{fig:ExpPower}
\end{figure}

After validating the RaccoonBot's electro-mechanical features, Fig.~\ref{fig:ExpPower} shows five different test trial profiles, tested with 25 minutes of separation between them. Those curves allow analyzing how the MPP changes through time, due to the position of the sun and the generated tree shades. In Fig~\ref{fig:ExpPower}, the red circles are the MPP highlighted at each power curve, that is also the robot's position in the wire were it ended after performing the solar tracking algorithm. Moreover, the black line connecting them traces the route the MPP follows through time, where the irregular path is mostly due to the irregular shadows from the trees, which makes the MPP impossible to track without the algorithm.

\vspace{-10pt}
\vspace{-7.5pt}

\begin{figure}[h!]
    \centering
    \includegraphics[scale=0.435]{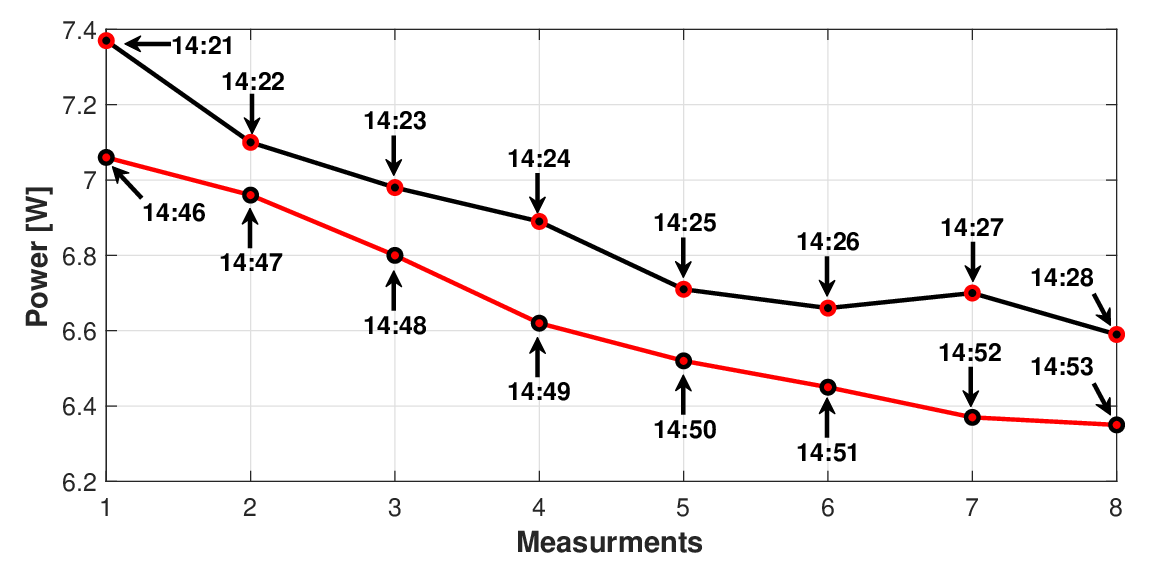}\vspace{-10pt}
    \caption{Power changing through time at a given $x_{Best}$ before the 14:28:00 (black line) and the 14:53:42 (red line) optimization trials from Fig.~\ref{fig:ExpPower}.}
    \label{fig:ExpPower_ch}
\end{figure}

Yet, Fig.~\ref{fig:ExpPower_ch} shows why is it necessary to continue analyzing power data when the robot settles in the MPP, since the profiles address how the power decreases through time when the robot is stationary in the MPP. Consequently, that it why the contributed algorithm also contemplates the reinitialization stage to find a new MPP to continue charging its battery with the best possible conditions.

\vspace{-10pt}
\vspace{-7.5pt}

\begin{figure}[h!]
    \centering
    \includegraphics[scale=0.435]{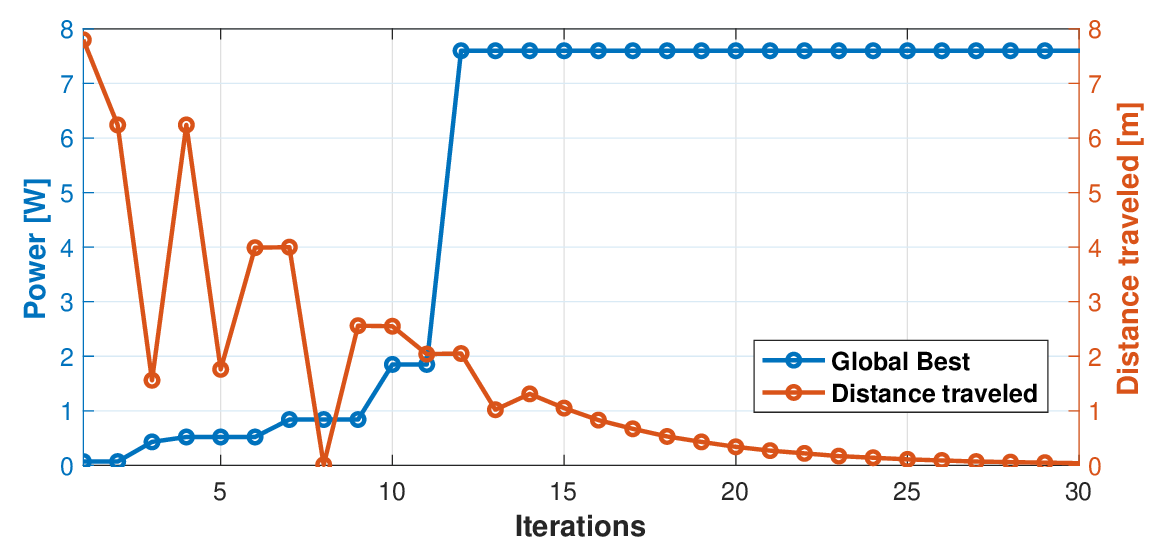}\vspace{-10pt}
    \caption{Convergence plot of $G_{Best}$ (left y-axis) and the traveled distance (right y-axis) during the 14:03 test trial.}
    \label{fig:convergence}
\end{figure}

Deepening into the reinitialization stages from the algorithm, Fig.~\ref{fig:convergence} shows how the robot through the algorithm behaved in the 14:03 optimization trial, where the algorithm was reinitialized with the initial conditions ($\Delta x = 7.8[m]$) and converges at the MPP (as addressed by the yellow curve from Fig.~\ref{fig:ExpPower}).

\vspace{-10pt}
\vspace{-7.5pt}

\begin{figure}[h!]
    \centering
    \includegraphics[scale=0.435]{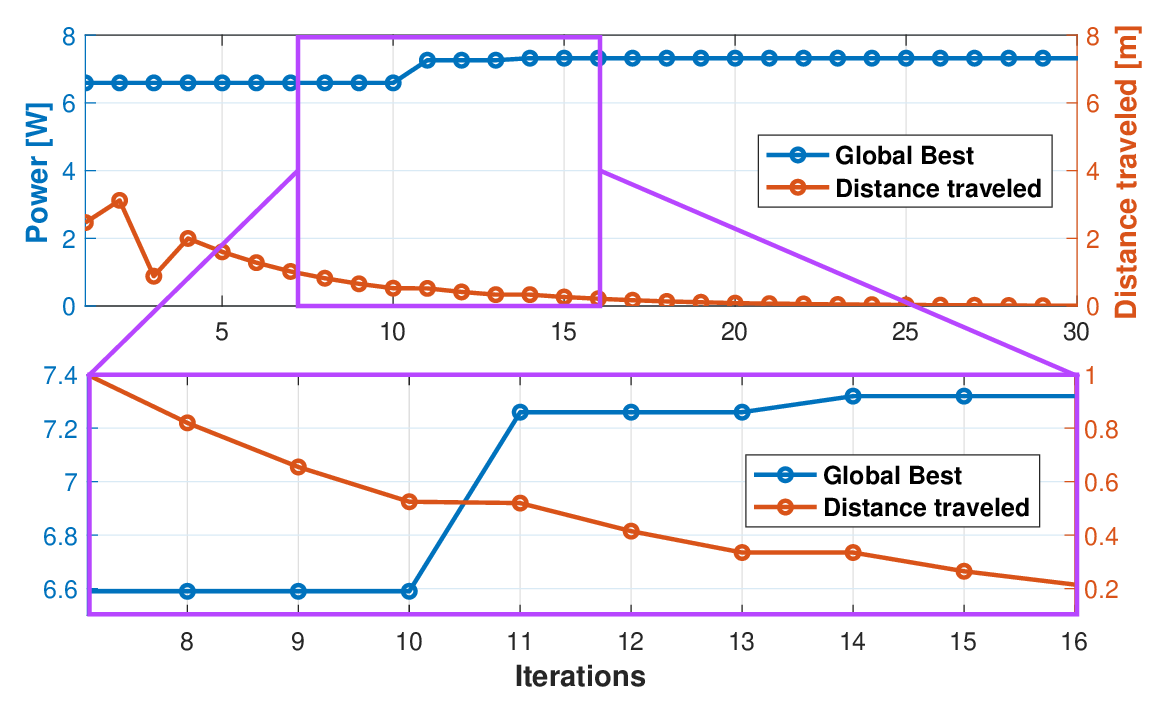}\vspace{-10pt}
    \caption{Convergence plot of $G_{Best}$ (left y-axis) and the traveled distance (right y-axis) during the 14:28 test trial.}
    \label{fig:convergence_fine}
\end{figure}

On the other hand, Fig.~\ref{fig:convergence_fine} presents the convergence plot from the 14:28 test trial, where the algorithm was reinitialized with the finer search conditions ($\Delta x = 2.25[m]$) and converges at a new MPP. The zoomed-in section, highlights how the finer search enables a more precise convergence into the MPP, and how the smaller steps also mean faster dynamics with lower power consumption; which from the power electronics perspective is a desirable feature since the energy harvested increases and the battery discharging rate decreases through time.

\vspace{-10pt}
\vspace{-10pt}

\section{Conclusions}\label{sec:Conclusions}
Environmental monitoring can be achieved through robotic platforms and in this paper, we introduced the RaccoonBot as a novel and reliable platform designed explicitly for this purpose. In particular, this work presents the integration of the electro-mechanical features that allow the robot to adapt to different wire-traversing scenarios, coupled with a fail-safe design that can even self-lock in the case of electrical shortage due to the non-reversibility properties of the proposed worm-gears mechanism. But, persistence is not only related to a fail-safe design but also to energetic autonomy. As such, this work also validated the algorithm in charge of the energy-aware mobility in charge of optimizing the energy harvested from PV power supply, meaning that the RaccoonBot represents a new design capable of fulfilling the requirements to act as a reliable platform for persistent environmental monitoring.








\balance 
\bibliographystyle{IEEEtran}
\bibliography{References.bib,NewRoboEcoBibRef.bib}

\end{document}